\documentclass[runningheads]{llncs}
\usepackage[T1]{fontenc}
\usepackage{graphicx}
\usepackage{amsmath,amsfonts}
\usepackage{algorithmic}

\usepackage{adjustbox}    % For resizing tables and figures
\usepackage[table]{xcolor} % For cell coloring (e.g., \cellcolor)
\usepackage{graphicx}     % For \rotatebox
\usepackage{booktabs}     % For better-looking tables (optional, for cleaner lines)

\newcommand{\argmax}{\operatornamewithlimits{argmax}}

\usepackage{multirow}
\usepackage{subcaption}
\usepackage{xspace}

\usepackage[capitalize,nameinlink,sort&compress]{cleveref}
\crefname{section}{Sec.}{Secs.}
\Crefname{section}{Section}{Sections}
\Crefname{table}{Table}{Tables}
\newcommand{\boldhline}{\specialrule{0.15em}{0em}{0.1em}}
\renewcommand{\arraystretch}{1.25}

\newcommand{\minus}[1]{\textcolor{red}{\textbf{- #1}}}
\newcommand{\plus}[1]{\textcolor{blue}{\textbf{+ #1}}}

\makeatletter
\DeclareRobustCommand\onedot{\futurelet\@let@token\@onedot}
\def\@onedot{\ifx\@let@token.\else.\null\fi\xspace}

\def\eg{\emph{e.g}\onedot} 
\def\ie{\emph{i.e}\onedot}

\def\etal{\emph{et al}\onedot}
\makeatother
\begin{document}
\title{MILE: Mixture of Incremental LoRA Experts for Continual Semantic Segmentation across Domains and Modalities}

% \author{Shishir Muralidhara \inst{1}\orcidID{0000-0001-7942-4698} \and
% Didier Stricker\inst{1,2}\orcidID{0009-0004-8794-6858} \and
% Ren\'e Schuster\inst{1,2}\orcidID{0000-0001-7055-9254}}

\author{Shishir Muralidhara \inst{1} \and
Didier Stricker\inst{1,2} \and
Ren\'e Schuster\inst{1,2}}
\authorrunning{Muralidhara et al.}
\titlerunning{MILE: Mixture of Incremental LoRA Experts}% Part of RIGHT running header

\institute{German Research Center for Artificial Intelligence (DFKI), \\ Trippstadter Str. 122, 67663 Kaiserslautern, Germany
\and
University of Kaiserslautern-Landau (RPTU), \\Gottlieb-Daimler-Str. 47, 67663 Kaiserslautern, Germany \\
\email{\{firstname.lastname\}@dfki.de}}
\maketitle              % typeset the header of the contribution
\begin{abstract}
Continual semantic segmentation requires models to adapt to new domains or modalities without sacrificing performance on previously learned tasks. 
Expert-based learning, in which task-specific modules specialize in different domains, has proven effective in mitigating forgetting. 
These methods include dynamic expansion, which suffers from scalability issues, or parameter isolation, which constrains the ability to learn new tasks.
We introduce Mixture of Incremental LoRA Experts (MILE), a modular and parameter-efficient framework for continual segmentation across both domains and modalities. 
MILE leverages Low-Rank Adaptation (LoRA) to instantiate lightweight experts for each new task while keeping the pretrained base network frozen. 
Each expert is trained exclusively on its task data, thus avoids overwriting previously learned information.
A prototype-guided gating mechanism dynamically selects the most appropriate expert at inference. 
MILE achieves the benefits of expert-based learning while overcoming its scalability limitations. 
It requires only a marginal parameter increase per task and tens of LoRA adapters are needed before matching the size of a single full model, making it highly efficient in both training and storage. 
Across domain- and modality-incremental benchmarks, MILE achieves strong performance while ensuring better stability, plasticity, and scalability.

\end{abstract}
\section{Introduction}
\label{sec:intro}

Semantic segmentation is essential for autonomous driving, as it provides precise pixel-level understanding of the environment and supports key perception tasks.
Despite significant progress driven by deep learning, most existing semantic segmentation models operate under rigid assumptions.
They are trained under closed-world settings with large labeled datasets from a fixed domain, assume a consistent set of input modalities, and are designed to solve a single task. 
In real-world scenarios, autonomous vehicles are continually exposed to a wide range of distributional shifts.
Domain shifts arise from environmental changes such as transitioning between different geographical locations, varying weather conditions, and lighting, all of which can drastically alter the visual appearance of the same semantic classes.
These shifts frequently result in performance degradation \cite{distribution_shift}, as static models fail to generalize beyond their training. 
Retraining or fine-tuning on new data leads to catastrophic forgetting \cite{catastrophic_forgetting}, where performance on previously learned domains or modalities is adversely affected.
Continual Learning (CL) has emerged as a promising paradigm to address the limitations of static models by enabling them to incrementally learn new tasks while preserving previously learned knowledge.
This introduces the stability-plasticity dilemma \cite{stability_plasticity}, a trade-off between learning new information (plasticity) and preserving prior knowledge (stability).
This challenge arises because updating model parameters to learn new tasks inevitably leads to overwriting previous weights resulting in forgetting.
In safety-critical applications such as autonomous driving, ensuring consistent and reliable performance is paramount, as any degradation in perception capabilities can lead to potentially hazardous outcomes.
An effective strategy to mitigate forgetting and the stability-plasticity trade-off is expert-based learning, where separate expert modules are assigned to a task. 
This approach minimizes interference between tasks by training each expert in isolation, preventing the overwriting of previously learned weights and preserving task-specific knowledge.
However, this approach suffers from poor scalability, as the model size increases linearly with the number of tasks or domains. \looseness-1
  
To address these challenges, we propose MILE: Mixture of Incremental LoRA Experts, a framework for continual semantic segmentation across domains and modalities.
MILE leverages Low-Rank Adaptation (LoRA) \cite{LoRA} to enable computationally efficient updates for adapting to new tasks.
It leverages the strengths of expert-based learning by using task-specific modules to prevent interference and forgetting, while addressing scalability limitations.
Instead of instantiating entire networks for each domain, MILE introduces compact LoRA-based expert modules that represent only a small fraction of the full model parameters.
This allows the system to scale to a large number of domains, requiring many such experts before even approaching the size of a single full model.
As new domains or modalities are encountered, MILE instantiates and trains a corresponding LoRA expert, while keeping the shared pretrained weights frozen. 
This significantly reduces the number of trainable parameters, thereby lowering the computational cost of training and storage.
During inference, we dynamically infer the domain of the input using a gating network and select the corresponding task expert. % for segmentation.

%-------------------------------------------------------------------
\section{Background and Related Works}
\label{sec:background}

Incremental learning involves training models on a sequence of tasks, enabling them to acquire new knowledge while retaining previously learned information. 
Depending on the objective of the task, two main incremental learning settings \cite{continual_semseg} are commonly considered:
Class-incremental learning, where the input distribution remains fixed but new, non-overlapping subsets of classes are introduced sequentially. 
This setting has been extensively explored in image classification \cite{cil_survey}, and more recently in semantic segmentation \cite{css_survey}.
In domain-incremental learning, the output space remains consistent, and the input distribution changes across tasks.
Modality-incremental learning \cite{MIL} extends domain-incremental learning to scenarios where new modalities are incrementally introduced.
Additionally, recent works propose novel incremental settings that address evolving ontologies \cite{LECO,CLEO} or simultaneous domain and class shifts \cite{LwS,versatile_il}, reflecting increasingly complex real-world learning challenges.

 \subsection{Domain-Incremental Learning}

The task of adapting to new domains has been widely explored in domain adaptation (DA) and domain generalization (DG) methods \cite{da_survey}.
DA methods assume access to both source and target domain data during adaptation to the target domain. 
In contrast, domain-incremental learning (DIL) involves sequentially learning new domains without access to data from previous domains.
Kalb \etal \cite{continual_semseg} evaluate the effectiveness of continual learning approaches for both class and domain incremental semantic segmentation, and observe replay-based methods are more effective for the latter. 
However, with replay-based approaches, there is an overhead for storing images for rehearsal or generating images and the subsequent pseudo-labeling.
MDIL \cite{MDIL} addresses incremental learning across geographic domains by combining shared domain-invariant parameters with domain-specific adapters and decoders, and uses the task-ID at inference to select the appropriate domain-specific components.
PSS \cite{PSS} incrementally adapts to adverse driving conditions through a dynamically growing collection of domain-specific expert networks, which are automatically selected at inference using a convolutional autoencoder ensemble.
Replaying Styles \cite{ReplayingStyles} uses low-level style embeddings to replay past domains without storing raw data.
RCIL \cite{RCIL} incrementally learns cities from Cityscapes \cite{CS} using two parallel branches to decouple old and new knowledge.

\subsection{Modality-Incremental Learning}

Modality-Incremental Learning (MIL) \cite{MIL} extends incremental learning to scenarios where data from new sensor modalities such as RGB, depth, infrared, or thermal are introduced sequentially.
Unlike domain-incremental learning, where the input distribution shifts within the same modality, MIL introduces drastically different input characteristics, making feature alignment and knowledge transfer more challenging. 
Similar to DIL, the label space remains consistent across tasks, with the same set of classes learned.
Hegde \etal \cite{MIL} propose DRMN, which uses modality-specific relevance maps to activate disjoint subsets of network parameters. 
Even under joint training where all modalities are available simultaneously, a single model struggles to learn effectively \cite{MIL}, highlighting the inherent difficulty of handling modality differences.
MILE simultaneously addresses shifts across both domains and modalities, enabling the model to adapt to new environments and sensor types within a unified framework.

\subsection{Expert-based Learning}

Expert-based approaches mitigate catastrophic forgetting by assigning dedicated sub-networks and networks \ie experts to individual tasks or domains. 
Expert learning can be achieved either dynamically, through network expansion, or within a fixed network through parameter isolation.
Dynamic network expansion adds new expert networks \cite{progressiveNNs,expertGate} as tasks arrive, enabling the model to expand its capacity over time.
Parameter isolation operates within a fixed network capacity by assigning task-specific parameters.
This can be achieved through masking \cite{piggyback} important weights for previous tasks, creating task-specific paths \cite{pathnet}, or pruning \cite{packnet} to free capacity for new tasks.
Notably, several recent continual semantic segmentation methods \cite{MDIL,PSS,MIL} leverage expert-based learning to achieve strong performance.
However, a critical limitation of expert-based approaches is scalability.
Dynamically growing networks increase model size and storage costs as new tasks are added, which can become prohibitive over long sequences.
Conversely, parameter isolation may struggle as the number of tasks increases, risking saturation of learning capacity and limiting adaptability.

%-------------------------------------------------------------------
\section{Mixture of Incremental LoRA Experts}
\label{sec:mile}
We propose Mixture of Incremental LoRA Experts (MILE), a framework that overcomes the scalability limitations of expert-based learning while retaining their benefits of task-specialization and strong knowledge preservation.
MILE builds on Low-Rank Adaptation (LoRA) \cite{LoRA} to enable modular and scalable adaptation. 
For each new domain or modality, a lightweight LoRA module is added while the base network and previous LoRA modules remain frozen. 
MILE does not require task-ID at inference and dynamically identifies the current domain using a lightweight gating network and selects the most appropriate LoRA expert for the given input.
MILE offers the following benefits:

\begin{itemize}
    \item \textbf{Parameter Efficiency and Scalability}. 
    MILE incrementally adapts to new domains by adding compact LoRA adapters that constitute only a small portion of the full model size, ensuring scalability to long task sequences.
    \item \textbf{Reduced Forgetting through Expert Isolation}. By confining adaptation to these small parameter subsets, MILE preserves previously learned information by preventing adverse interference, thereby avoiding forgetting.
    \item \textbf{Stability and Plasticity:} MILE overcomes the trade-off by isolating past knowledge in independent LoRA experts while adding new ones for full plasticity, ensuring strong performance on both old and new tasks. 
    \item \textbf{Resource-Efficient Adaptation:} By training only a small subset of parameters, MILE reduces both computational and memory overhead, making it well-suited for continual learning in resource-constrained environments.
    \item \textbf{Unified Learning of Domains and Modalities:} Through a modular mixture-of-experts approach, MILE supports both domain and modality-incremental learning within a single architecture. 
\end{itemize}

\begin{figure*}[t]
	\centering
	\includegraphics[width=\linewidth]{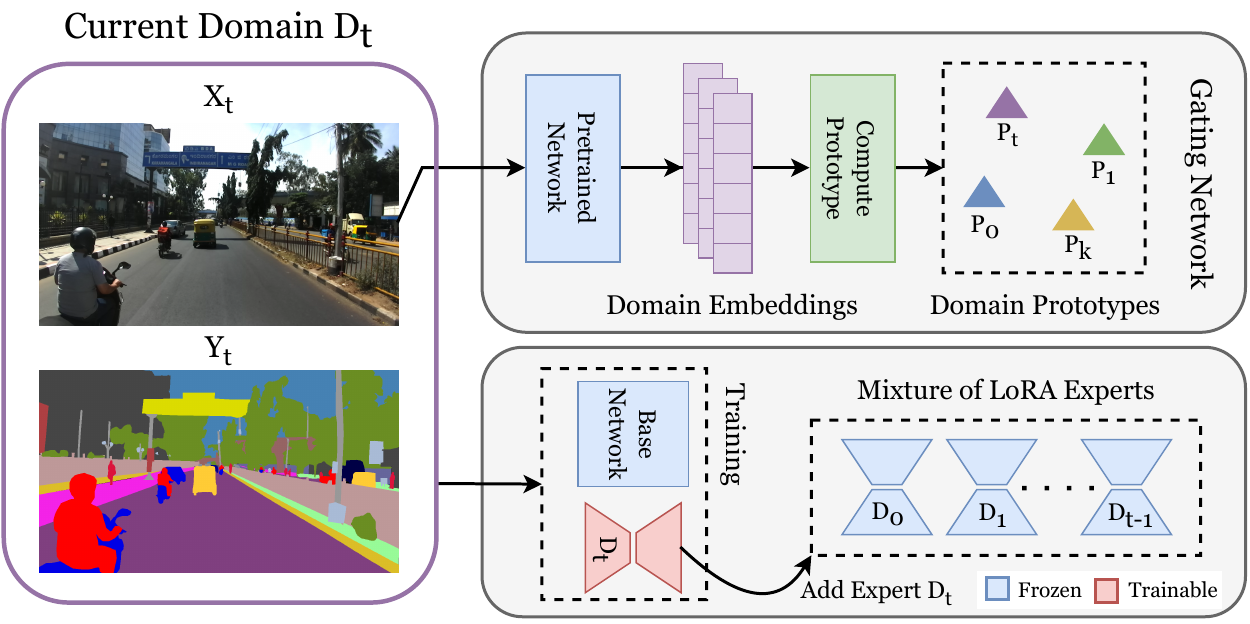}
	\caption{Overview of the training process in MILE. Using the input images $X_t$, the domain prototype $P_t$ is computed using a frozen pretrained network. LoRA expert $D_t$ is instantiated and trained while keeping the base network frozen.}
	\label{fig:mile_training}
\end{figure*}

\subsection{Training with Low-Rank Adaptation}
Incremental learning involves a model sequentially learning from a series of tasks $T = \{t_1, t_2, \ldots, t_n\}$ while retaining knowledge from previous tasks. 
Each task $t$ is associated with task-specific data $D_t = (X_t, Y_t)$. 
In both domain- and modality-incremental settings, the input distribution shifts between tasks ($X_{t-1} \neq X_t$), but the number of classes $C$ in $Y$ remains consistent across all tasks.
Adapting to new tasks typically necessitates retraining the entire network, which is computationally prohibitive in resource-constrained environments. 

In contrast, our approach leverages Low-Rank Adaptation \cite{LoRA} for parameter-efficient continual learning \cite{CLoRA}. 
LoRA is a parameter-efficient fine-tuning (PEFT) method that adapts large pre-trained models to downstream tasks with minimal computational cost.
For a network with weights $W \in \mathbb{R}^{d \times k}$, LoRA introduces a small, low-rank update $\Delta W$, represented as the product of two matrices: $B \in \mathbb{R}^{d \times r}$ and $A \in \mathbb{R}^{r \times k}$, where $r \ll d$ determines the number of trainable parameters. 
During training, only the LoRA weights are updated, while the base network weights $W$ remain frozen. 
This significantly reduces both the computational and memory requirements, making LoRA particularly suitable for continual learning in resource-constrained settings.

We use LoRA to incrementally adapt to new domains or modalities. 
An overview of the training process with MILE is presented in \cref{fig:mile_training}.
For each new task $t$, we train only the corresponding LoRA weights, denoted as $\Delta W_t$. 
The full model for any task can be reconstructed by combining the frozen base network weights with the appropriate task-specific LoRA weights. 
This modular design requires storing only LoRA weights, enabling efficient scaling with a large number of tasks and addressing the scalability concerns of expert-based learning. 
It would take tens of LoRA modules before their combined size approaches that of a full model, highlighting the storage efficiency of our approach.
Despite updating only a subset of parameters through LoRA, MILE achieves performance comparable to full network training, while significantly improving computational efficiency.

\begin{figure*}[t]
    \centering
    \includegraphics[width=1\linewidth]{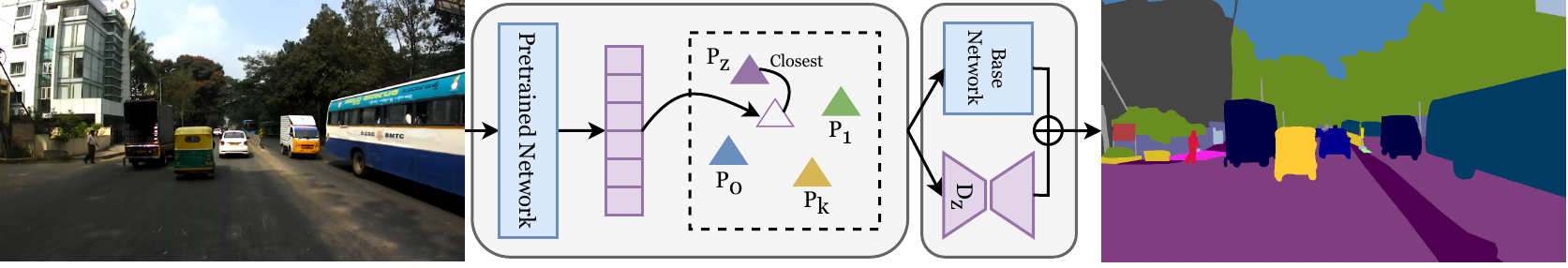}
    \caption{Overview of inference in MILE. For an image, domain features are extracted using the frozen network, and the gating network infers the domain $z$ by finding the closest prototype and then selects the expert $D_z$ for segmentation.}
    \label{fig:mile_inference}
\end{figure*}

\subsection{Inference with Prototype-Based Gating}

Our approach trains task-specific LoRA experts for each domain or modality, and a key challenge at inference is selecting the appropriate expert without reintroducing catastrophic forgetting at the domain inference stage. 
In incremental learning, where experts are added incrementally, it is not possible to jointly train a classifier to route inputs to the appropriate expert.
Updating such a classifier with new tasks can inadvertently overwrite knowledge from previously learned domains.
To overcome this limitation, we propose a prototype-based gating mechanism that avoids the need for a trainable classifier. 
 
For each task $t$, we compute a domain prototype $\mathbf{p}_t$ as the mean of the features of all training samples $X_t \in D_t$:
\begin{equation}
    \mathbf{p}_t = \frac{1}{|D_t|} \sum_{(x, y) \in D_t} f_\theta(x),
\end{equation}
where $f_\theta(\cdot)$ denotes a frozen feature extractor.
$\mathbf{p}_t$ serves as a compact representation of the overall domain characteristics.  
During inference, given an input sample $x$, we first extract its feature representation:
$\mathbf{z} = f_\theta(x).$
The gating function $g (\mathbf{z})$ selects the expert corresponding to the prototype most similar to $\mathbf{z}$:
\begin{equation}
    g(\mathbf{z}) = \argmax_{t \in \{1, \ldots, n\}} \text{cosine\_sim}(\mathbf{z}, \mathbf{p}_t)
\end{equation}
where $n$ is the total number of tasks. 
 
An overview of inference in MILE with the prototype-based gating network is presented in \cref{fig:mile_inference}.
This approach offers several advantages: It enables domain identification without requiring a trainable classifier, eliminating the need for additional model updates.
It scales efficiently with an increasing number of tasks, as it only requires storing a single prototype per task, ensuring minimal memory overhead.
Additionally, it inherently mitigates forgetting, since the feature extractor remains frozen throughout the process. 
The fixed feature space ensures consistent performance as new tasks are added without interfering with or altering the representations of previously learned tasks.

%-------------------------------------------------------------------
\section{Experiments and Results}
\label{sec:results}

In this section, we describe the datasets and task settings, drawing from prior work on domain- and modality-incremental learning.
We evaluate MILE under realistic challenges, including adverse weather, geographic shifts, and modality changes, and report results with a focus on scalability and efficiency.
\subsection{Datasets}

We evaluate MILE on diverse segmentation datasets capturing domain and modality shifts. 
Cityscapes (CS) \cite{CS} provides urban street scenes under clear weather with 19 classes, while ACDC \cite{ACDC} and Berkeley DeepDrive (BDD) \cite{BDD} introduce domain variations due to adverse weather, illumination, and geographic diversity using the same label set.
Indian Driving Dataset (IDD) \cite{IDD} includes challenging conditions such as unstructured environments with 26 classes, including domain-specific labels. 
Freiburg Thermal \cite{FT} introduces a modality shift between aligned RGB and thermal infrared image pairs with 13 classes.

\subsection{Implementation and Baselines}

For semantic segmentation, we use SegFormer-B5 \cite{Segformer} from SATS \cite{SATS} for all methods. 
MILE uses LoRA with rank $r{=}16$ which corresponds to 5\% of the total model parameters, enabling parameter-efficient adaptation. 
We use batch size 12 and the learning rates are 0.01/0.001 for standard training and 0.05/0.005 for LoRA during initial/incremental steps. 
The gating network uses a pretrained ConvNeXt-Tiny \cite{ConvNeXt} for feature extraction.
% We compare against standard continual learning baselines: Single-Task (\textbf{ST}), Fine-Tuning (\textbf{FT}), Joint Training (JT), 
We evaluate MILE against standard continual learning baselines: Single-Task (ST) models are trained independently on each task and serve as a reference to measure forgetting. 
Fine-Tuning (FT) adapts the model sequentially to new tasks, resulting in positive forward transfer but significant forgetting of previous tasks. 
Joint Training (JT) trains a single model on all task data simultaneously, typically providing an upper-bound performance.
Elastic Weight Consolidation (EWC) \cite{EWC} constrains model updates to preserve parameters important for previous tasks.
Incremental Learning Techniques (ILT) \cite{ILT} uses knowledge distillation to retain knowledge across tasks.
We additionally report comparisons with PSS \cite{PSS}, MDIL \cite{MDIL}, and DRMN \cite{MIL} using results from their original papers. 
All methods are evaluated using performance change $\triangle P$ relative to corresponding single-task models.

\begin{table*}[t]
\centering
\caption{
Results (mIoU) on the sequence $CS_{day} \rightarrow ACDC_{fog} \rightarrow ACDC_{rain} \rightarrow ACDC_{snow} \rightarrow ACDC_{night}$ after learning all tasks.}
\begin{adjustbox}{width=\linewidth}
\begin{tabular}{c||cc||cc||cc||cc||cc||c} 
\boldhline
\textbf{Method} & \textbf{$CS$} & \textbf{$\triangle P$ (\%)}  & \textbf{$Fog$} &  \textbf{$\triangle P$ (\%)}  & \textbf{$Rain$} &  \textbf{$\triangle P$ (\%)}  & \textbf{$Snow$} &  \textbf{$\triangle P$ (\%)}  & \textbf{$Night$} &  \textbf{$\triangle P$ (\%)}  & \textbf{Average}  \\
\boldhline
ST / PSS$^\dagger$ & 71.79 & - & 60.05 & - & 57.83 & - & 60.03 & - & 47.85 & - & 59.51 \\ \boldhline

Fine-Tune & 63.26 & \minus{11.88} & 65.43 & \plus{08.96} & 58.22 & \plus{00.67} & 56.41 & \minus{06.03} & 50.93 & \plus{06.44} & 58.85 \\ \hline

EWC & 65.48 & \minus{08.79} & 68.13 & \plus{13.46} & 57.87 & \plus{00.07} & 55.99 & \minus{06.73} & 48.42 & \plus{01.19} & 59.18 \\ \hline

ILT & 66.33 & \minus{07.61} & 65.89 & \plus{09.73} & 58.04 & \plus{00.36} & 55.73 & \minus{07.16} & 39.19 & \minus{18.10} & 57.04 \\ \boldhline

MILE$^\dagger$ & 71.21 & \minus{00.81} & 70.55 & \plus{17.49} & 64.17 & \plus{10.96} & 66.20 & \plus{10.28} & 49.95 & \plus{04.39} & 64.42 \\ \hline

MILE \phantom{ } & 71.21 & \minus{00.81} & 70.30 & \plus{17.07} & 64.06 & \plus{10.77} & 65.16 & \plus{08.55} & 49.95 & \plus{04.39} & 64.14 \\ \boldhline

Joint-Train & 72.29 & \plus{00.70} & 75.65 & \plus{25.98} & 67.05 & \plus{15.94} & 67.07 & \plus{11.73} & 52.19 & \plus{09.07} & 66.85 \\ \boldhline
\multicolumn{12}{l}{\footnotesize{$\dagger$ denotes domain inference using an oracle.}} \\
\end{tabular}
\end{adjustbox}
\label{tab:adv_domains}
\end{table*}

\subsection{Adverse Weather Domains}

Using Cityscapes (CS) \cite{CS} and ACDC \cite{ACDC} we formulate a fine-grained categorization of adverse weather conditions with the following task sequence: $CS_{day} \rightarrow ACDC_{fog} \rightarrow ACDC_{rain} \rightarrow ACDC_{snow} \rightarrow ACDC_{night}$, consisting of 5 tasks. 
PSS \cite{PSS} is evaluated on a broader categorization between normal and adverse weather conditions grouped together.
We use the results from the fully trained single-task models to represent PSS evaluated with an oracle during inference.
The results after learning all tasks are presented in \cref{tab:adv_domains}.
A limitation with single-task models is the lack of positive forward transfer when learning new tasks. 
When learning tasks with small incremental datasets, such as the individual domains from ACDC, all sequential learning methods demonstrate noticeable improvements.
Constraining parameter updates with EWC \cite{EWC} and ILT \cite{ILT}, hinders learning on new tasks and, although less severe than fine-tuning, still leads to partial forgetting of earlier tasks.
MILE$^\dagger$ with an oracle to provide domain-ID consistently improves performance across all ACDC domains compared to single-task models, and comparable results on the initial task. 
When using a gating network, MILE achieves results close to the oracle, with only a slight drop in performance due to misrouted samples.

\subsection{Geographical Domains}

Using Cityscapes (CS) \cite{CS}, BDD \cite{BDD}, and IDD \cite{IDD}, we evaluate the task sequence $CS \rightarrow BDD \rightarrow IDD$, where each task represents a domain shift in the geographical location.
To enable sequential learning, we use IDD with labels mapped to the 19 CS classes. 
The task-wise results after learning a new domain and evaluating on all previously seen domains are presented in \cref{tab:geo_domains}.
In contrast to the results presented in \cref{tab:adv_domains}, we can observe that the sequentially trained models do not exhibit any positive forward transfer. 
Even with the fine-tuning, where learning on the new tasks is not regularized or constrained, the results on the new domain \ie $BDD$ in $Step \ 2$ and $IDD$ in $Step \ 3$ fall short of the single-task models.
With EWC \cite{EWC} and ILT \cite{ILT}, we observe results marginally better than fine-tuning, highlighting the difficulty of using a single model in incrementally learning new domains.
The joint training model, which had previously improved performance across all domains, underperforms compared to the single-task models in this setting, presumably due to substantial differences in domain characteristics across continents.
Our proposed approach MILE, achieves results closest to the single-task models.

\begin{table*}[t]
\centering
\caption{Step-wise results (mIoU) for the task sequence $CS \rightarrow BDD \rightarrow IDD$ representing different geographical locations.}
\begin{adjustbox}{width=\linewidth}

\begin{tabular}{c||c||cc|cc||cc|cc|cc||c} 

\boldhline
\multirow{2}{*}{\textbf{Method}} 
& \multicolumn{1}{c||}{\textbf{$Step \  1: CS$}} 
& \multicolumn{4}{c||}{\textbf{$Step \  2: CS \rightarrow BDD$}} 
& \multicolumn{6}{c||}{\textbf{$Step \  3: CS \rightarrow BDD \rightarrow IDD$}} 
& \multirow{2}{*}{\textbf{Average}} \\ \cline{2-12}
& \textbf{$CS$} 
& \textbf{$CS$} & \textbf{$\triangle P$ (\%)} 
& \textbf{$BDD$} & \textbf{$\triangle P$ (\%)} 
& \textbf{$CS$} & \textbf{$\triangle P$ (\%)} 
& \textbf{$BDD$} & \textbf{$\triangle P$ (\%)} 
& \textbf{$IDD$} & \textbf{$\triangle P$ (\%)} 
&  \\ 
\boldhline

ST / PSS$^\dagger$
& 71.79 & 71.79 & - & 60.28 & - & 71.79 & - & 60.28 & - & 74.10 & - & 68.72 \\ \boldhline

Fine-Tune 
& 71.79 
& 63.35 & \minus{11.76} & 60.22 &  \minus{00.10}
& 55.53 & \minus{22.65} & 50.22 & \minus{16.69} & 73.54 & \minus{00.76}
& 59.76 \\ \hline

EWC 
& 71.79 
& 67.81 & \minus{05.54} & 57.64 & \minus{04.38}
& 59.76 & \minus{16.76} & 52.30 & \minus{13.24} & 67.52 & \minus{08.88}
& 59.86 \\ \hline

ILT 
& 71.79 
& 68.94 & \minus{03.97} & 55.58 & \minus{07.80}
& 64.22 & \minus{10.54} & 53.79 & \minus{10.77} & 64.28 & \minus{13.25}
& 60.76 \\ \boldhline

MILE$^\dagger$ 
& 71.21 
& 71.21 & \minus{00.81} & 59.12 & \minus{01.92}
& 71.21 & \minus{00.81} & 59.12 & \minus{01.92}  & 72.37 & \minus{02.33}
& 67.57 \\ \hline

MILE
& 71.21 
& 71.21 & \minus{00.81} & 58.99 & \minus{02.14}
& 71.21 & \minus{00.81} & 58.35 & \minus{03.20} & 71.05 & \minus{04.12}
& 66.87 \\ \boldhline

Joint-Train 
& 66.35
& 66.35 & \minus{07.58} & 56.24 &  \minus{06.70}
& 66.35 & \minus{07.58} & 56.24 &  \minus{06.70} & 71.32 & \minus{03.75}
& 64.64 \\ \boldhline

\multicolumn{13}{l}{\footnotesize{$\dagger$ denotes domain inference using an oracle.}} \\

\end{tabular}
\end{adjustbox}
\label{tab:geo_domains}
\end{table*}

\begin{table*}[b]
\centering
\caption{Step-wise results (mIoU) for the task sequence $RGB \rightarrow IR \rightarrow Gray$ representing modality shifts.}
\begin{adjustbox}{width=\linewidth}

\begin{tabular}{c||c||cc|cc||cc|cc|cc||c} 

\boldhline
\multirow{2}{*}{\textbf{Method}} 
& \multicolumn{1}{c||}{\textbf{$Step \  1: RGB$}} 
& \multicolumn{4}{c||}{\textbf{$Step \  2: RGB \rightarrow IR$}} 
& \multicolumn{6}{c||}{\textbf{$Step \  3: RGB \rightarrow IR \rightarrow Gray$}} 
& \multirow{2}{*}{\textbf{Average}} \\ \cline{2-12}
& \textbf{$RGB$} 
& \textbf{$RGB$} & \textbf{$\triangle P$ (\%)} 
& \textbf{$IR$} & \textbf{$\triangle P$ (\%)} 
& \textbf{$RGB$} & \textbf{$\triangle P$ (\%)} 
& \textbf{$IR$} & \textbf{$\triangle P$ (\%)} 
& \textbf{$Gray$} & \textbf{$\triangle P$ (\%)} 
&  \\ 
\boldhline

ST / PSS
& 80.16 & 80.16 & - & 63.03 & - & 80.16 & - & 63.03 & - & 78.51  & - & 73.90 \\ \boldhline

Fine-Tune 
& 80.16
& 10.29 & \minus{87.16} & 56.98 & \minus{09.60} 
& 78.50 & \minus{02.07} & 06.68 & \minus{89.40} & 77.99 & \minus{00.66} 
& 54.39 \\ \hline

EWC 
& 80.16
& 66.22 & \minus{17.39} & 49.71 & \minus{21.13} 
& 78.18 & \minus{02.47} & 05.68 & \minus{90.99} & 76.90 & \minus{02.05} 
& 53.59 \\ \hline

ILT 
& 80.16
& 09.91 & \minus{87.64} & 43.14 & \minus{31.56} 
& 63.53 & \minus{20.75} & 08.68 & \minus{86.23} & 62.28 & \minus{20.67} 
& 44.83\\ \boldhline

MILE$^\wedge$ 
& 78.51 
& 78.51 & \minus{02.06} & 58.50 & \minus{07.19}  
& 78.51 & \minus{02.06} & 58.50 & \minus{07.19} & 76.46 & \minus{02.61}
& 71.16 \\ \hline

MILE$^\diamond$
&  78.51
&  78.51 & \minus{02.06}  & 60.76 & \minus{03.60} 
&  78.51 & \minus{02.06}  & 60.76 & \minus{03.60} & 76.94 & \minus{02.00}
&  72.07 \\ \boldhline

Joint-Train 
& 78.64
& 78.64 & \minus{01.90} & 23.62 & \minus{62.53} 
& 78.64 & \minus{01.90} & 23.62 & \minus{62.53} & 77.16 & \minus{01.72}
& 59.81 \\ \boldhline

\multicolumn{13}{l}{\footnotesize{$\wedge$ denotes sequentially trained models, $\diamond$ denotes models trained independently.}} \\

\end{tabular}
\end{adjustbox}
\label{tab:mod_domains}
\end{table*}

\subsection{Modality-Incremental}

The results on the Freiburg Thermal \cite{FT} dataset highlight the challenges posed by the significant distribution shift across modalities.
We evaluate using the task sequence $RGB \rightarrow IR \rightarrow Gray$ and present the results in \cref{tab:mod_domains}.
Unlike DIL, this setting does not require domain inference, as the task-ID that corresponds to the sensor is inherently known.
As expected, the large modality shifts present a significant challenge. 
Even the joint training model, with access to all modalities and no forgetting, underperforms, highlighting that a single model struggles to capture all modality-specific features.
Single-task models achieve the highest performance, indicating that isolating modalities allows the model to learn modality-specific characteristics without interference, demonstrating task-specialization, one of the advantages of expert-based learning.
The stability-plasticity trade-off in EWC and ILT, becomes even more challenging under large shifts.
We study the influence of task sequence order under additional experiments to understand how modality sequences affect forward transfer and forgetting.
Within the MILE framework, we observe that sequentially trained $MILE^\wedge$ underperforms compared to single-task $MILE^\diamond$, which is trained independently on each modality. 
These results highlight the challenges of modality-incremental learning and the limitations of joint training and sequential adaptation.
 
Additionally, we compare our method against RMN and DRMN from \cite{MIL}, using the relative performance $\triangle P$ to the corresponding single-task models. 
The results are summarized in \cref{tab:mod_comparison}. 
With parameter isolation in RMN and DRMN, the network capacity to learn new tasks becomes exhausted, affecting performance on later modalities. 
In contrast, MILE leverages the full network representation through individual LoRA modules, achieving performance closest to that of the corresponding single-task models across all modalities.

\begin{table}[tb]
\centering
\caption{Final results for the task $RGB \rightarrow IR \rightarrow Gray$.}
\begin{adjustbox}{width=0.8\linewidth}
\begin{tabular}{c||cc|cc|cc||c} 

\boldhline
\textbf{Method} 
& \textbf{RGB} & \textbf{$\triangle P$ (\%)} 
& \textbf{IR} & \textbf{$\triangle P$ (\%)} 
& \textbf{Gray} & \textbf{$\triangle P$ (\%)} & 
\textbf{Average} \\ 
\boldhline

Single-Task
& 80.16 & - & 63.03 & - & 78.51 & - & 73.90 \\ \hline

MILE
& 78.51 & \minus{02.06} & 60.76 & \minus{03.60} & 76.94 & \minus{02.00} & 72.07 \\ \boldhline

Single-Task \cite{MIL}
& 76.41 & - & 59.56 & - & 74.56  & - & 70.18 \\ \hline

RMN \cite{MIL}
& 73.13 & \minus{04.29} & 55.01 & \minus{07.64} & 68.29 & \minus{08.41}
& 65.48 \\ \hline

DRMN \cite{MIL}
& 73.21 & \minus{04.19} & 54.95 & \minus{07.74} & 69.38 & \minus{06.95}
& 65.85 \\ \boldhline

\end{tabular}
\end{adjustbox}
\label{tab:mod_comparison}
\end{table}

\section{Additional Experiments}
We present additional experiments, including the influence of task sequence in modality-incremental learning, domain incremental learning with heterogeneous labels, domain inference results, and an analysis of the scalability of MILE. %, and qualitative visualizations.

\begin{table*}[t]
\centering
\caption{Step-wise results (mIoU) for the task sequence $RGB \rightarrow Gray \rightarrow IR$ representing modality shifts.}
\begin{adjustbox}{width=\linewidth}
\begin{tabular}{c||c||cc|cc||cc|cc|cc||c} 

\boldhline
\multirow{2}{*}{\textbf{Method}} 
& \multicolumn{1}{c||}{\textbf{$Step \  1: RGB$}} 
& \multicolumn{4}{c||}{\textbf{$Step \  2: RGB \rightarrow Gray$}} 
& \multicolumn{6}{c||}{\textbf{$Step \  3: RGB \rightarrow Gray \rightarrow IR$}} 
& \multirow{2}{*}{\textbf{Average}} \\ \cline{2-12}
& \textbf{$RGB$} 
& \textbf{$RGB$} & \textbf{$\triangle P$ (\%)} 
& \textbf{$Gray$} & \textbf{$\triangle P$ (\%)} 
& \textbf{$RGB$} & \textbf{$\triangle P$ (\%)} 
& \textbf{$Gray$} & \textbf{$\triangle P$ (\%)} 
& \textbf{$IR$} & \textbf{$\triangle P$ (\%)} 
&  \\ 
\boldhline

ST / PSS
& 80.16 & 80.16 & - & 78.51 & - & 80.16 & - & 78.51 & - & 63.03 & - & 76.47 \\ \boldhline

Fine-Tune 
& 80.16
& 78.71 & \minus{01.81} & 78.29 & \minus{00.28}
& 21.42 & \minus{73.28} & 20.73 & \minus{73.60} & 59.19 & \minus{06.09}
& 33.78 \\ \hline

EWC 
& 80.16
& 78.23 & \minus{02.41} & 76.74 & \minus{02.25}
& 66.47 & \minus{07.08} & 61.71 & \minus{21.40} & 50.05 & \minus{20.59} 
& 59.41 \\ \hline

ILT 
& 80.16
& 79.35 & \minus{01.01} & 75.23 & \minus{04.18}
& 17.75 & \minus{77.86} & 16.26 & \minus{79.29} & 43.34 & \minus{31.24}
& 25.78 \\ \boldhline

MILE$^\diamond$
&  78.51
&  78.51 & \minus{02.06}  & 76.94 & \minus{02.00} 
&  78.51 & \minus{02.06}  & 76.94 & \minus{02.00} & 60.76 & \minus{03.60}
&  72.07 \\ \boldhline

Joint-Train 
& 78.64
& 78.64 & \minus{01.90} & 77.16 & \minus{01.72} 
& 78.64 & \minus{01.90} & 77.16 & \minus{01.72} & 23.62 & \minus{62.53}
& 59.81 \\ \boldhline

\end{tabular}
\end{adjustbox}
\label{tab:mod_domains_2}
\end{table*}

\subsection{Influence of Task Sequence}

The order in which tasks are presented plays a key role in balancing stability and plasticity between previously learned and new tasks.
Previously in \cref{tab:mod_domains}, we observed the performance on the first modality learned improve after learning the last task due to their visual similarity. 
Here, we evaluate using the sequence $RGB \rightarrow Gray \rightarrow IR$ and the results are presented in \cref{tab:mod_domains_2}.
Single-task, joint training and MILE trained independently are not affected by the task order. 
For sequentially trained methods, in Step 2, where $Gray$ is learned after $RGB$, the performance on the first modality remains relatively stable.
This suggests that the shared visual characteristics between $RGB$ and $Gray$ mitigate forgetting.
This behaviour contrasts with the earlier sequence $RGB \rightarrow IR \rightarrow Gray$ in \cref{tab:mod_domains}, where the second task $IR$ affects the performance on $RGB$ due to larger domain dissimilarity.
The addition of $IR$, which differs significantly from the previous modalities, adversely interferes with the previous tasks and results in substantial forgetting.
Consequently, the average performance in \cref{tab:mod_domains_2} across all modalities is lower than the previous task sequence $RGB \rightarrow IR \rightarrow Gray$.

\subsection{Domain-Incremental Learning with Heterogeneous Labels}

\begin{figure*}[t]
	\centering
	\begin{adjustbox}{width=\linewidth}
		\setlength{\tabcolsep}{1pt} 
		\renewcommand{\arraystretch}{0.5}
		
		\begin{tabular}{c c c c} 
			& \textbf{Cityscapes} & \textbf{BDD} & \textbf{IDD} \\
			
			\rotatebox{90}{Image} & 
			\includegraphics[width=0.3\textwidth]{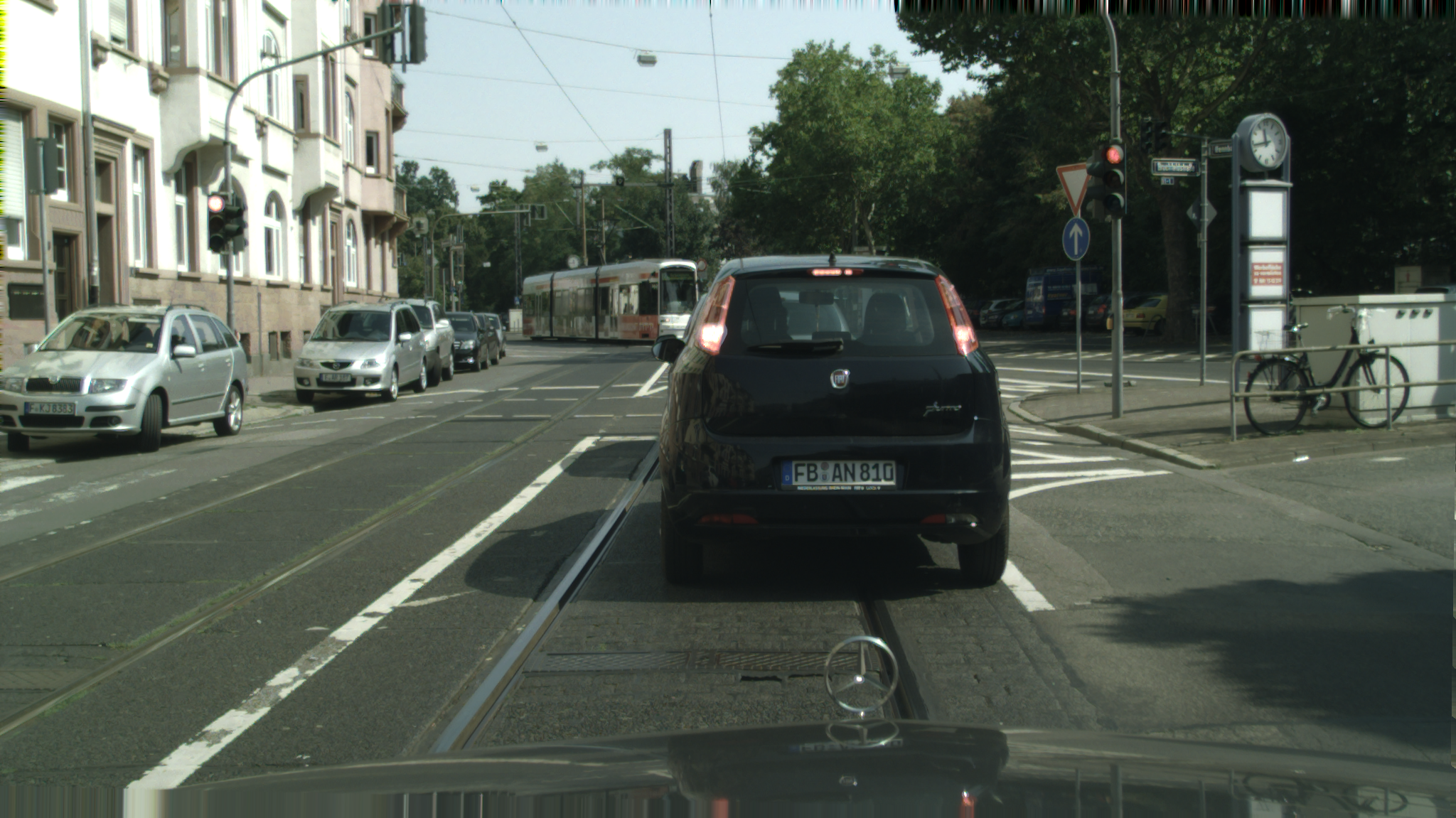} & 
			\includegraphics[width=0.3\textwidth]{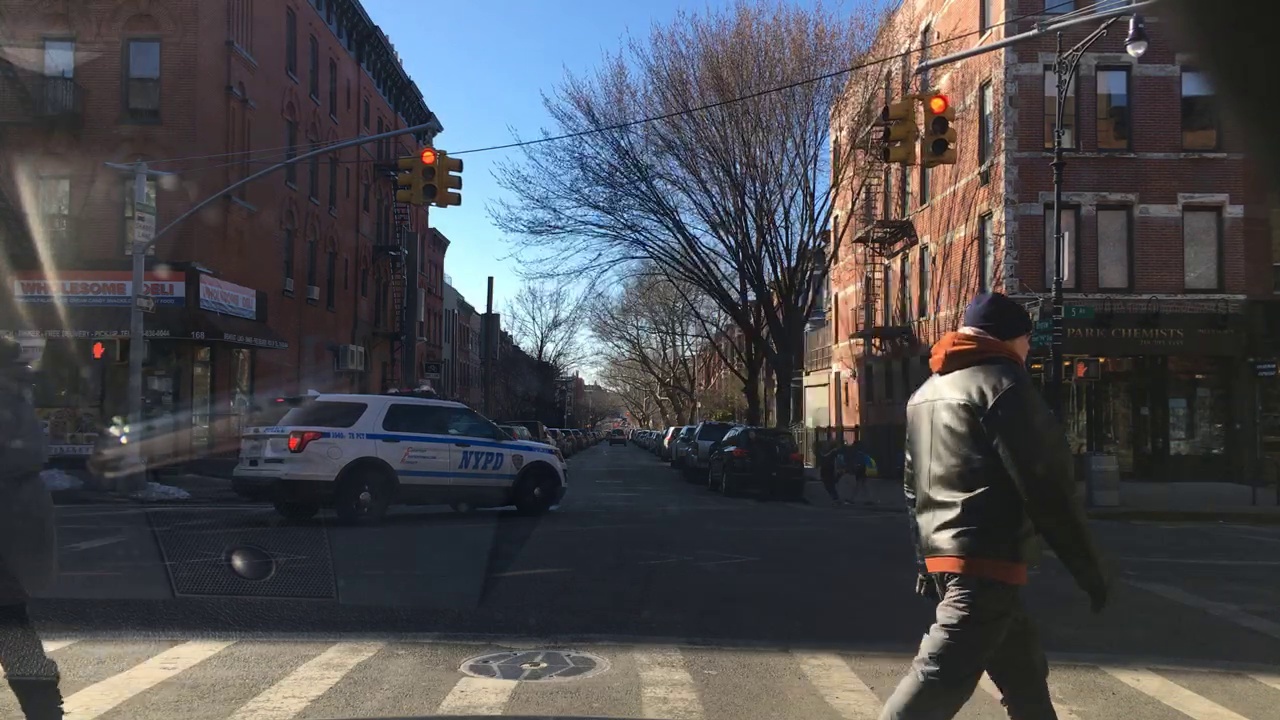} & 
			\includegraphics[width=0.3\textwidth]{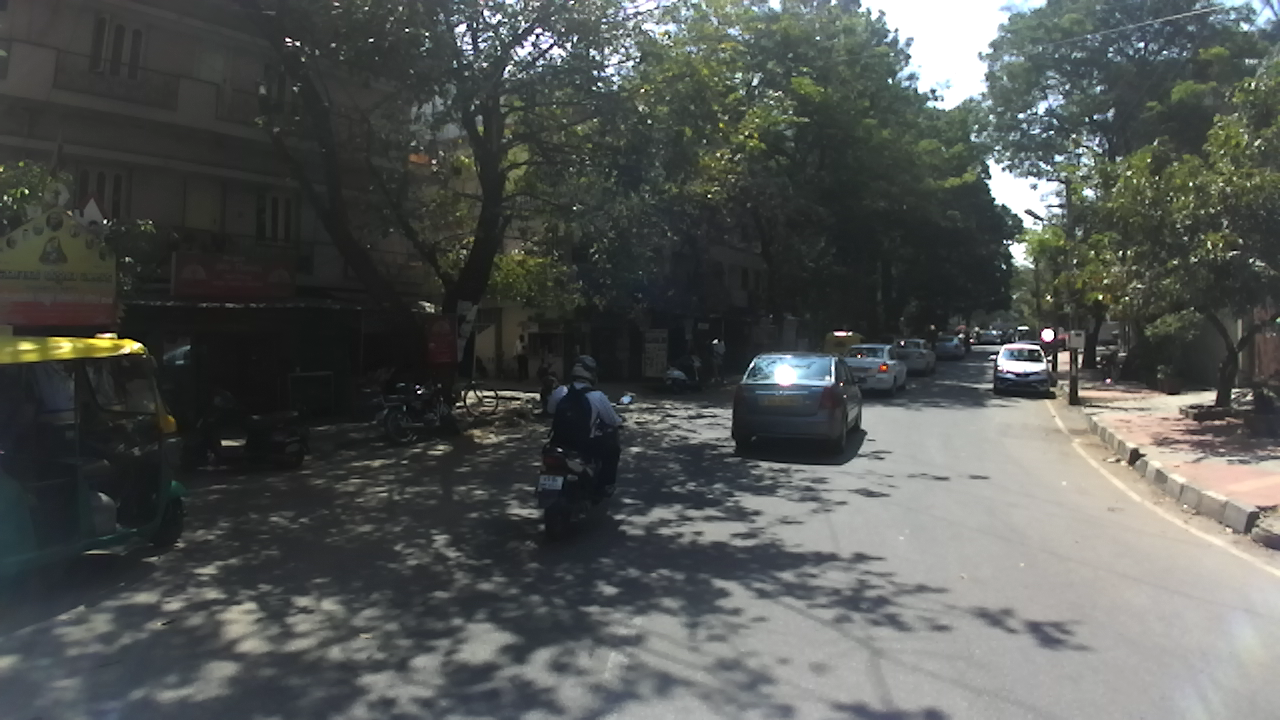} \\
			
			\rotatebox{90}{GT} & 
			\includegraphics[width=0.3\textwidth]{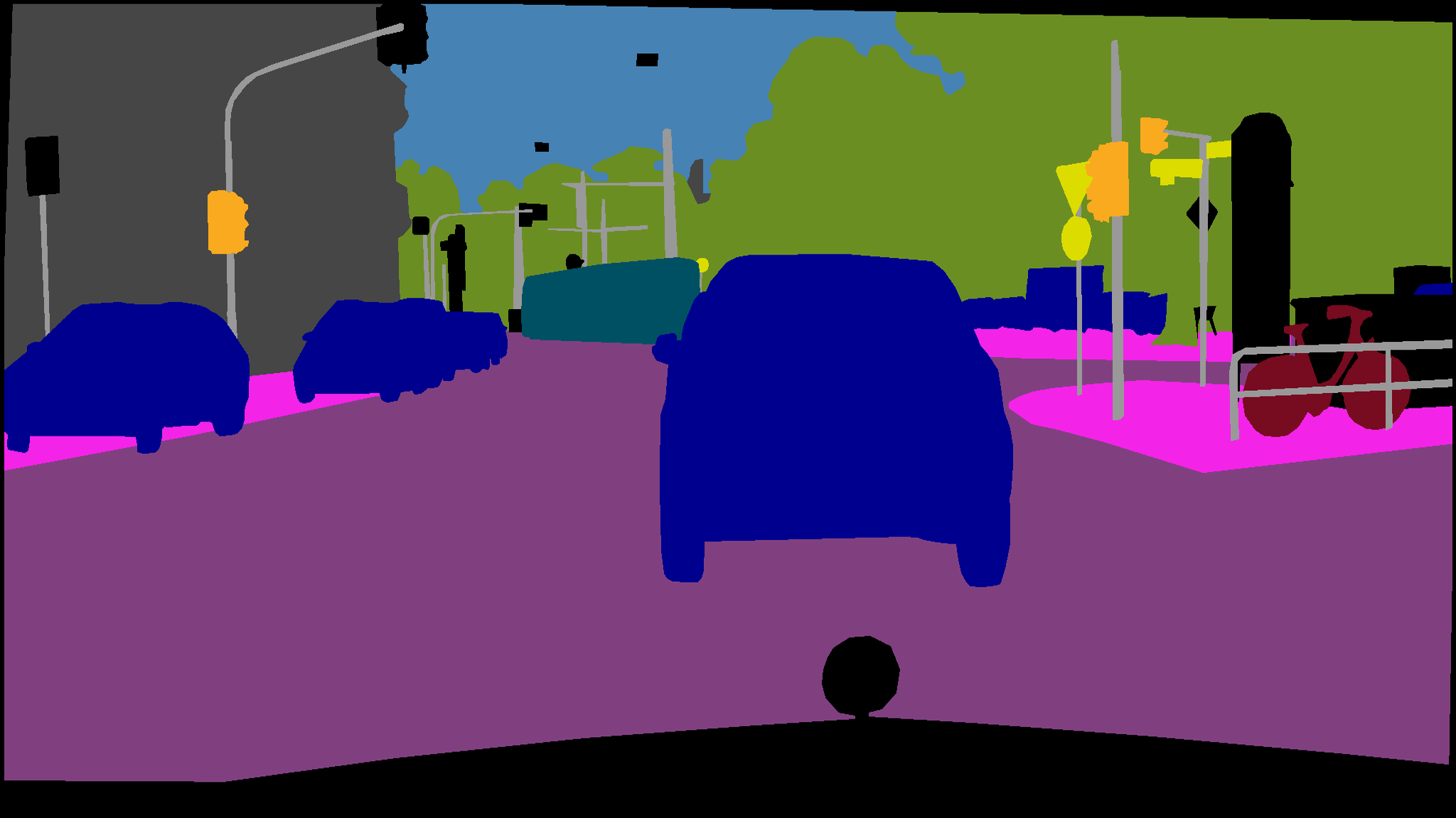} & 
			\includegraphics[width=0.3\textwidth]{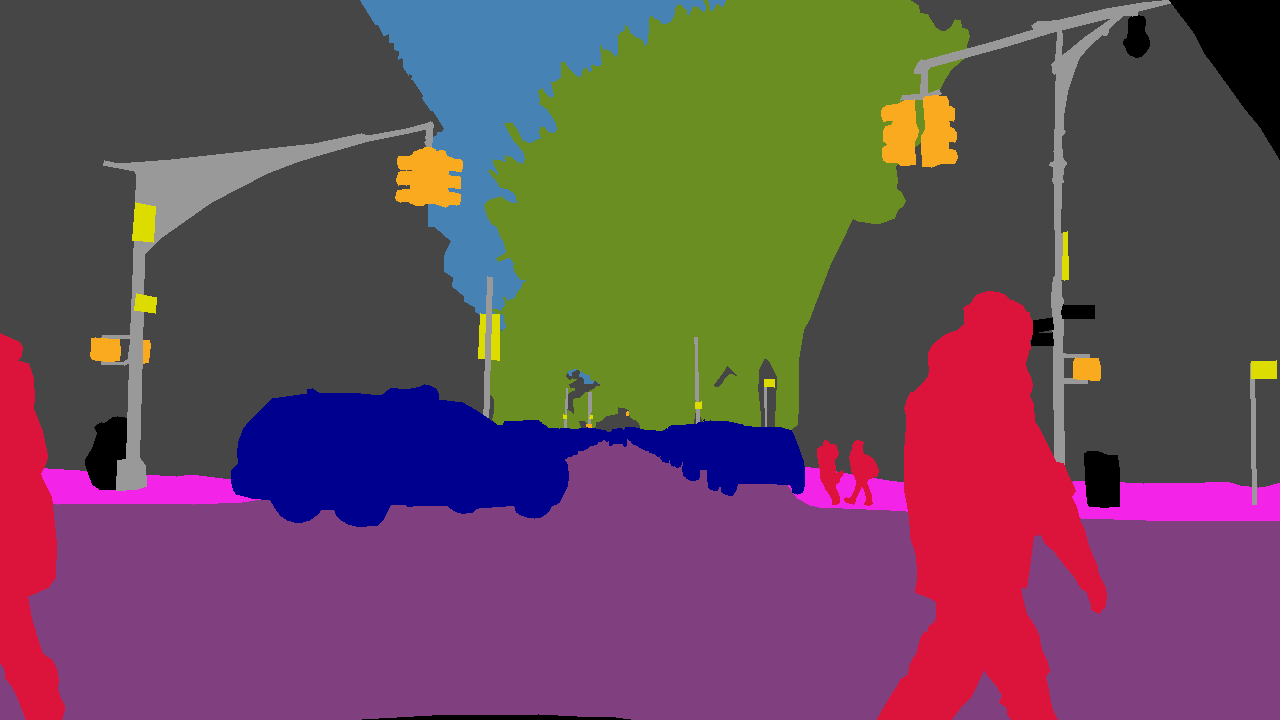} & 
			\includegraphics[width=0.3\textwidth]{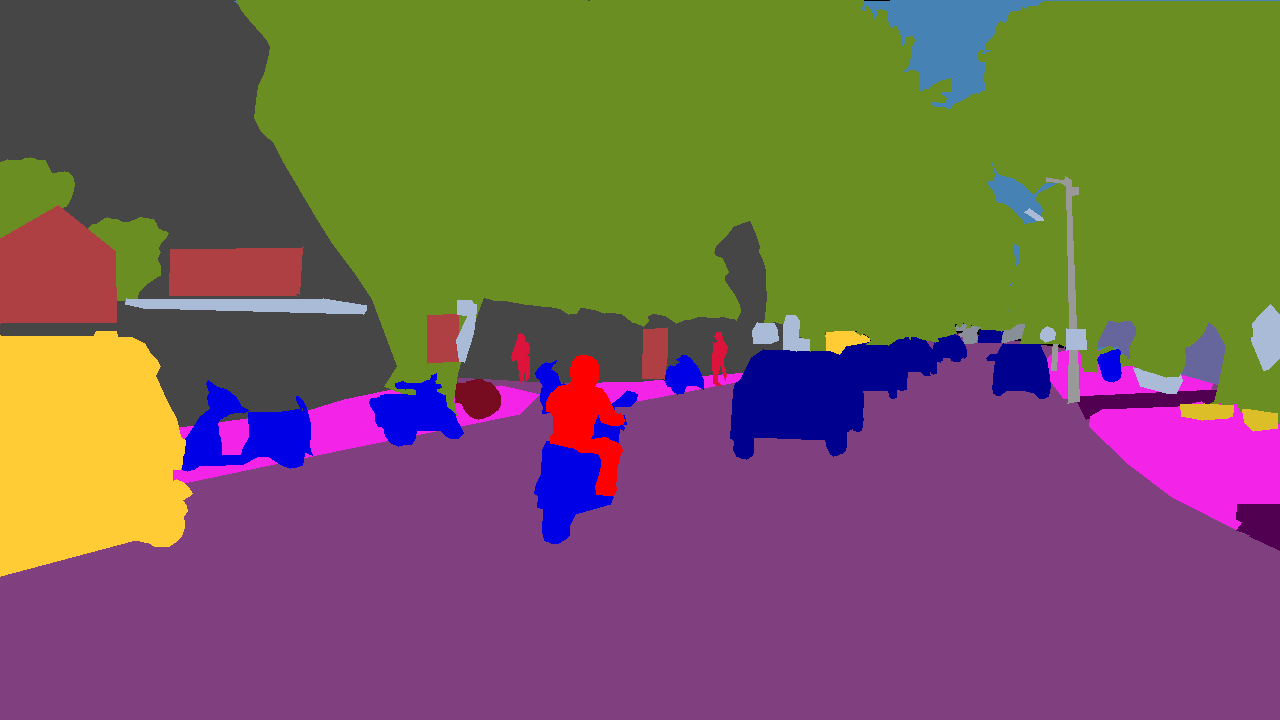} \\
			
			\rotatebox{90}{MILE} & 
			\includegraphics[width=0.3\textwidth]{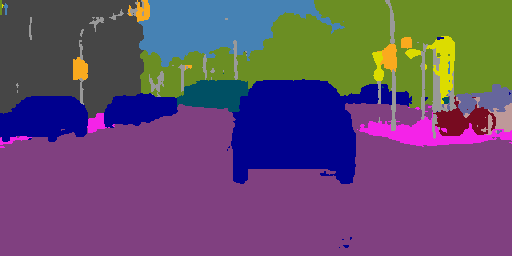} & 
			\includegraphics[width=0.3\textwidth]{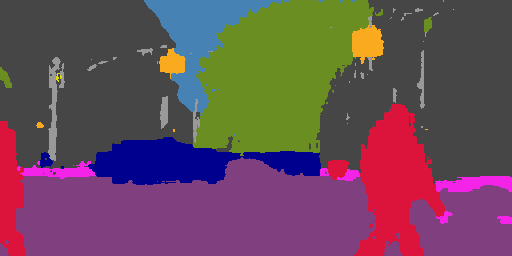} & 
			\includegraphics[width=0.3\textwidth]{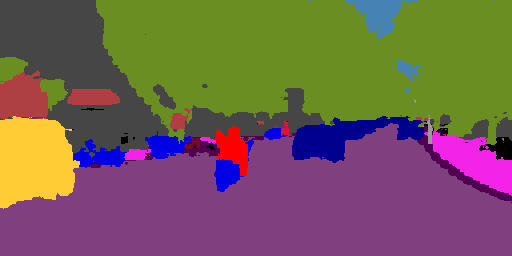} \\
			
			\rotatebox{90}{JT} & 
			\includegraphics[width=0.3\textwidth]{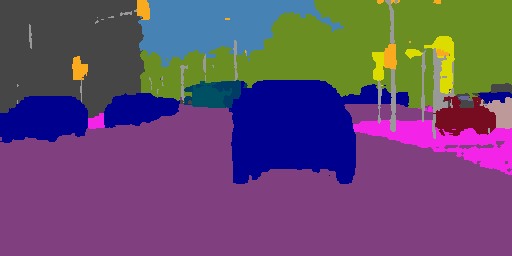} & 
			\includegraphics[width=0.3\textwidth]{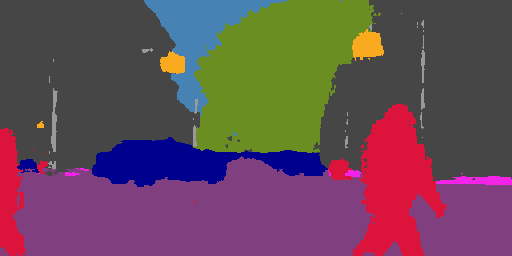} & 
			\includegraphics[width=0.3\textwidth]{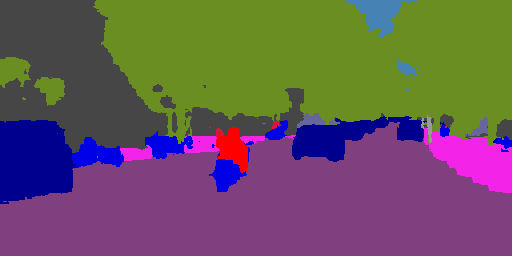} \\
			
		\end{tabular}
	\end{adjustbox}
	\caption{Qualitative comparison of segmentation results for domain-incremental learning of geographical domains using Cityscapes \cite{CS}, BDD \cite{BDD}, and IDD \cite{IDD}.}
	\label{fig:qual_results}
\end{figure*}

\begin{table}[b] 
    \centering 
    \caption{Final results (mIoU) for the task sequence $CS \rightarrow BDD \rightarrow IDD$.} 
    \begin{adjustbox}{width=0.85\linewidth} 
    \begin{tabular}{c||cc|cc|cc||c} 
        \boldhline 
        \textbf{Method} & \textbf{CS} & \textbf{$\triangle P$ (\%)} & \textbf{BDD} & \textbf{$\triangle P$ (\%)} & \textbf{IDD} & \textbf{$\triangle P$ (\%)} & \textbf{Average} \\ 
        \boldhline 
        Single-Task & 71.79 & - & 60.28 & - & 65.54 & - & 65.87 \\ 
        \hline 
        MILE$^\dagger$ & 71.21 & \minus{00.81} & 59.12 & \minus{01.92} & 64.85 & \minus{01.05} & 65.06 \\ 
        %\hline
        %MILE &  &  &  &  &  &  &  \\ 
        \boldhline 
        Single-Task \cite{MDIL} & 72.55 & - & 54.10 & - & 61.97 & - & 62.87 \\ 
        \hline 
        MDIL$^\dagger$ \cite{MDIL} & 59.19 & \minus{18.41} & 49.66 & \minus{08.21} & 59.16 & \minus{04.53} & 56.00 \\ 
        \boldhline 
        \multicolumn{8}{l}{\footnotesize{$\dagger$ denotes domain inference using an oracle.}} \\

    \end{tabular} 
    \end{adjustbox} 
    \label{tab:geo_hetero} 
\end{table}
In classical domain-incremental learning, the set of classes $C$ between domains remains consistent with $C_{t-1} = C_t$, and the model is only required to adapt to distributional shifts in the input space.
However, in many real-world scenarios, the label spaces across domains are not homogeneous.
For instance, the IDD dataset \cite{IDD} contains 26 classes, including domain-specific labels like auto-rickshaw, which have no direct correspondence in BDD \cite{BDD} or Cityscapes \cite{CS}. 
This illustrates how heterogeneous labels impose additional complexity: The model must expand its semantic space while still retaining representations of overlapping classes (\eg car, person, or traffic light) that are present across domains.
One common strategy to mitigate this issue is to map heterogeneous labels back into a common label space as done in \cref{tab:geo_domains} but this discards domain-specific and fine-grained information and undermines the utility of the model in that domain.  
Expert-based learning approaches with dedicated decoders \cite{MDIL} or separate networks remain unaffected by label heterogeneity, as each expert is trained independently and evaluated only on its corresponding label set.
The results on the task-sequence $CS \rightarrow BDD \rightarrow IDD$ with 26 classes in IDD is presented in \cref{tab:geo_hetero} and compared with MDIL \cite{MDIL}.
Beyond quantitative evaluation, \cref{fig:qual_results} presents qualitative results. 
Unlike joint training, which is constrained by a shared label space, MILE preserves domain-specific information, retaining unique classes such as auto-rickshaw in IDD.
\begin{figure}[t]
    \centering
    \begin{minipage}{0.48\textwidth}
        \centering
        \includegraphics[width=\linewidth]{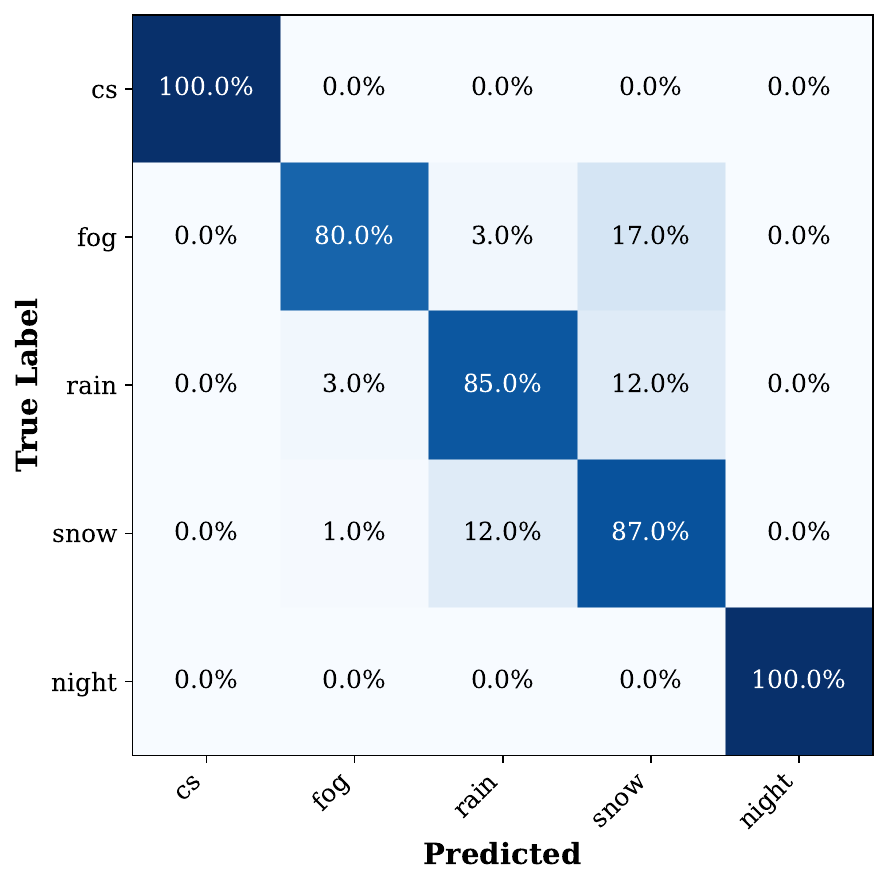}
        \caption{Normalized confusion matrix for domain inference illustrating the effectiveness of the gating network.}
        \label{fig:domain_inference}
    \end{minipage}
    \hfill
    \begin{minipage}{0.48\textwidth}
        \centering
        \includegraphics[width=\linewidth]{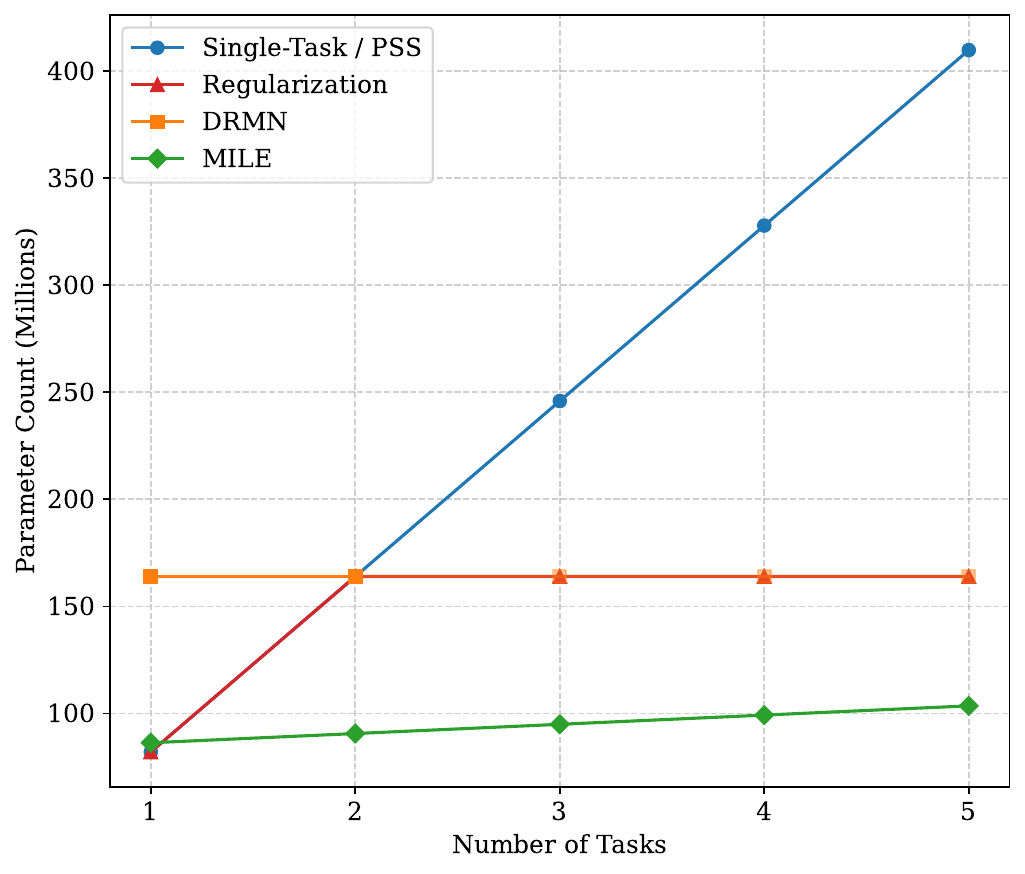}
        \caption{Parameter growth across tasks for different continual learning approaches. 
            MILE adds only a small number of LoRA parameters per task ensuring scalability.}
        \label{fig:scalability}
    \end{minipage}
\end{figure}

\subsection{Domain Inference}

We present the results of our domain inference stage, where the gating network identifies the domain and routes inputs to the corresponding task expert.
The normalized confusion matrices for the classification results on the validation sets are presented in \cref{fig:domain_inference}.
Even in the challenging multi-domain classification across five weather domains, our approach achieves good performance. 
The overall performance of MILE using this gating network for domain inference is close to MILE with an oracle for domain inference, highlighting the effectiveness of routing to the correct task expert.
The main advantage of our approach is that it does not require any additional training for domain inference, as we only use domain prototypes computed using frozen features. 

\subsection{Scalability in MILE}

\Cref{fig:scalability} illustrates the total parameter count increase with the number of tasks for different continual learning approaches.
Single-task and PSS \cite{PSS} grow linearly, as each task has a dedicated model, which limits scalability.
Regularization-based methods, maintain a constant parameter count across tasks, and only require the previous task model for regularizing the training of incremental tasks, but this reduces plasticity and can limit performance.
DRMN \cite{MIL} has a higher initial parameter count due to the relevance maps but its fixed network capacity can constrain learning over long task sequences.

MILE achieves a balance between these extremes: Using LoRA, MILE adds only a small number of parameters with each new task. 
Notably, even with several LoRA modules added for multiple tasks, the total parameter count remains well below that of a full task-specific model.
This demonstrates a trade-off in continual learning approaches: Expert-learning based methods are highly effective but suffer from poor scalability, whereas regularization methods are scalable but are less flexible. 
MILE addresses these competing objectives, retaining the advantages of task-specialization while limiting parameter growth as new tasks are added, making it a practical solution for scalable and efficient continual learning.

\section{Conclusion}
In this work, we propose MILE (Mixture of Incremental LoRA Experts), a modular and parameter-efficient framework for continual semantic segmentation across domains and modalities. 
MILE uses LoRA to add lightweight task-specific experts while keeping the base network frozen, preventing knowledge overwriting. 
A prototype-based gating mechanism selects the appropriate expert at inference.
We extensively evaluate MILE across diverse domain-incremental settings, including adverse weather, geographical domains, and modalities. 
Across these tasks, MILE achieves performance on par with upper-bound baseline, demonstrating its effectiveness in balancing scalability, efficiency, and performance.

% \clearpage

\section*{Acknowledgments}
This work was partially funded by the German Federal Ministry of Research, Technology, and Space under the project COPPER (16IW24009).

{
    \small
    \bibliographystyle{splncs04}
    \bibliography{main}
}

\end{document}